\documentclass[letterpaper, 10pt, conference]{ieeeconf}  

\usepackage{graphicx}
\usepackage{subfigure}
\usepackage{amsfonts}
\IEEEoverridecommandlockouts                              

\title{\LARGE \bf
Adaptive Tuning of Robotic Polishing Skills based on \\Force Feedback Model
}

\author{Yu Wang$^{1}$, Zhouyi Zheng$^{1}$, Chen Chen$^{2*}$, Zezheng Wang$^{1}$, Zhitao Gao$^{1}$ 
\\Fangyu Peng$^{3}$, Xiaowei Tang$^{1}$, and Rong Yan$^{1}$ 
\thanks{*Corresponding author.
        {\tt\small chenchen$\_$1014@foxmail.com}}%
\thanks{$^{1}$Yu Wang, Zhouyi Zheng, Zezheng Wang, Zhitao Gao, Xiaowei Tang and Rong Yan are with School of Mechanical Science and Engineering, Huazhong University of Science and Technology, China.}%
\thanks{$^{2}$Chen Chen is with Hubei Key Laboratory of Mechanical Transmission and Manufacturing Engineering, Wuhan University of Science and Technology, China.}%
\thanks{$^{3}$Fangyu Peng is with State Key Laboratory of Digital Manufacturing Equipment and Technology, School of Mechanical Science and Engineering, Huazhong University of Science and Technology, China.}%
}

\begin{document}

\maketitle
\thispagestyle{empty}
\pagestyle{empty}

\begin{abstract}

Acquiring human skills offers an efficient approach to tackle complex task planning challenges. When performing a learned skill model for a continuous contact task, such as robot polishing in an uncertain environment, the robot needs to be able to adaptively modify the skill model to suit the environment and perform the desired task. The environmental perturbation of the polishing task is mainly reflected in the variation of contact force. Therefore, adjusting the task skill model by providing feedback on the contact force deviation is an effective way to meet the task requirements. In this study, a phase-modulated diagonal recurrent neural network (PMDRNN) is proposed for force feedback model learning in the robotic polishing task. The contact between the tool and the workpiece in the polishing task can be considered a dynamic system. In comparison to the existing feedforward neural network phase-modulated neural network (PMNN), PMDRNN combines the diagonal recurrent network structure with the phase-modulated neural network layer to improve the learning performance of the feedback model for dynamic systems. Specifically, data from real-world robot polishing experiments are used to learn the feedback model. PMDRNN demonstrates a significant reduction in the training error of the feedback model when compared to PMNN. Building upon this, the combination of PMDRNN and dynamic movement primitives (DMPs) can be used for real-time adjustment of skills for polishing tasks and effectively improve the robustness of the task skill model. Finally, real-world robotic polishing experiments are conducted to demonstrate the effectiveness of the approach.

\end{abstract}

\section{INTRODUCTION}

 Learning from demonstration (LfD) is increasingly being used for robotic contact tasks, including tactile tasks \cite{RN327}\cite{RN330}\cite{9739464}, assembly tasks \cite{RN328}\cite{RN329}\cite{8961854}, cutting \cite{RN331}\cite{RN332}, writing \cite{RN333} and polishing \cite{RN334}\cite{RN335}. Traditional kinematics-based skill learning often overlooks vital force and stiffness information, making it challenging to apply effectively to intricate tasks involving complex force interactions \cite{10011841}\cite{10011996}. Among these tasks, contact force tracking is particularly important, especially in continuous contact tasks such as robot polishing. However, in contact tasks with environmental uncertainty, when the measured force signal deviates from expected values, it is possible to try to correct the skill action by correlating the force information with the movement primitive. Force signals measured by force sensors are the basis of this work. Mapping the contact force error to the adjustment amount of the skills model is one of the ways to address this problem \cite{RN327}, which can enhance the robustness of the contact task skills model. Initially, some hand-designed feedback models \cite{RN357}\cite{RN358} were used to map the sensor-measured error to the amount of movement primitive correction. However, these models were limited to specific tasks with non-high-dimensional sensing. Therefore, some data-driven approaches \cite{RN349} were proposed to generalize the learning of sensor feedback models.

However, in tasks like polishing, in addition to the relative sliding between the tool and the workpiece, there is also rotation of the tool or workpiece, and various factors influence the relationship between the amount of trajectory correction and the contact force. In robotic polishing tasks, the tool-to-workpiece contact force is related to the physical parameters of the tool and workpiece, the contact depth and area, and the relative rotational speed \cite{RN260}. The mapping relationship between force error and skill model adjustment in the feedback model of polishing task  serves the same main purpose as force tracking control methods such as impedance control. Impedance control adjusts acceleration, velocity and position based on contact force errors, and thus, the two processes influence each other during force tracking. So the relationship between contact force error and skill model adjustment is analogous to a dynamic system. Moreover, robot polishing skills consist of time-series data, where data at different moments can have an influence on each other. And Recurrent neural network (RNN) can memorize historical information to adapt to more complex dynamic environments \cite{WOS:000564561700002}. So RNN is more suitable for force sensor feedback model building compared to feedforward neural network (FNN). It is worth noting that the polishing contact process can be modeled to be a dynamic system, and constructing a sensor feedback learning model that is more suitable for dynamic systems is more suitable for the generalization task of polishing skills. To better model the force error feedback, we propose a novel force sensor feedback learning model called PMDRNN. PMDRNN takes into account the interactive dynamics between the tool and the workpiece, as well as the effect of adjacent moments, and the effect of the phase function in the skill model DMP, which satisfies the learning of dynamical system features associated with the canonical system in DMPs. By combining the phase-modulated term\cite{RN327} with the diagonal recurrent neural network (DRNN) \cite{RN337} to improve the accuracy of the force sensor feedback learning model.

\begin{figure*}[htbp]
  \centering
  \includegraphics[width=12cm]{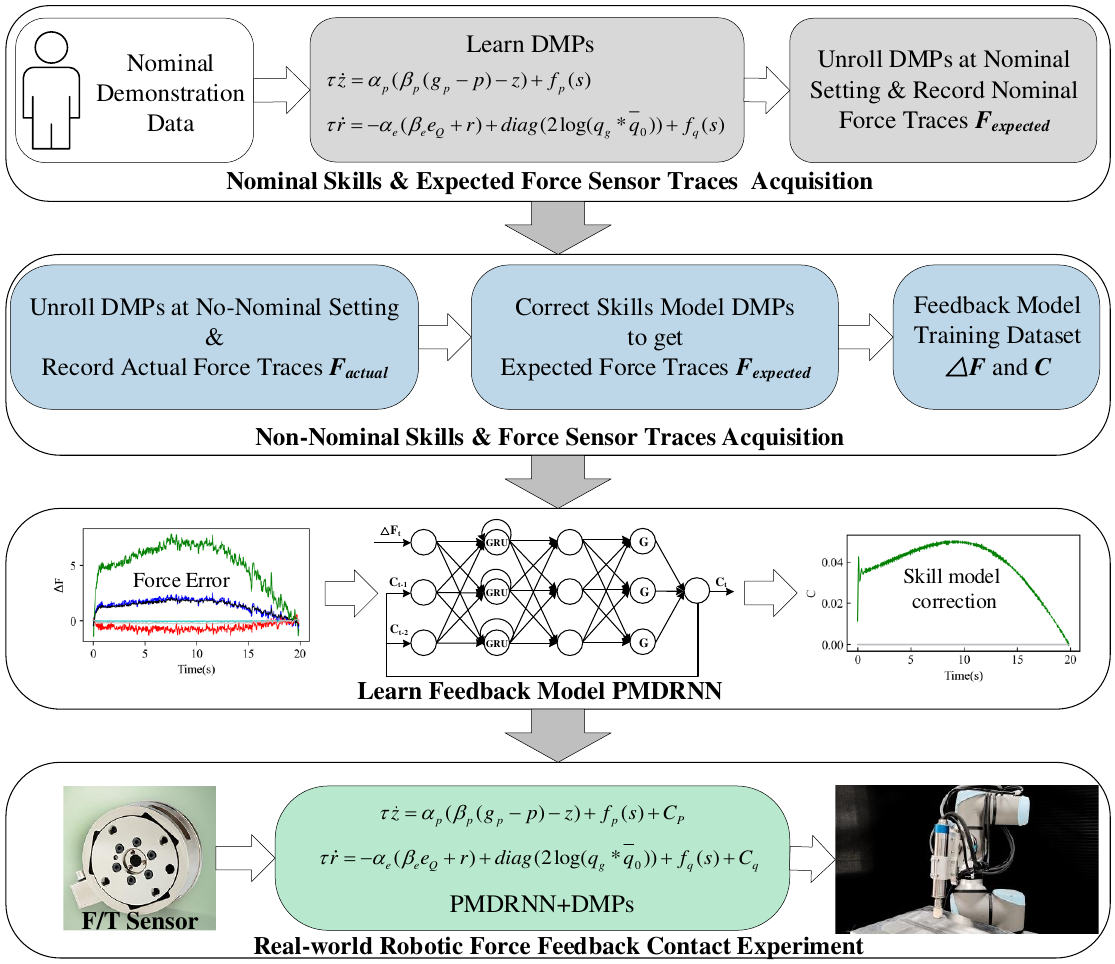}
  \caption{This figure shows the process of using the proposed method for learning and implementing feedback strategies. The force feedback model is first trained based on data from real-world robotic demonstration tasks, and then the trained model is used to adapt the task skill model in real-time to achieve the desired contact skills}
  \label{Figure1}
\end{figure*}

Combining the trained PMDRNN with the DMPs allows for real-time modification of the skill model while considering the force feedback error. This approach differs from directly modifying the forcing term weights and residual terms in the DMPs model, which is a characteristic of contact tasks. The main process of this study is illustrated in Fig.\ref{Figure1}. Firstly, the operator constructs the desired skill model DMPs and the desired contact force trajectories by demonstrating the nominal skills. Secondly, the actual contact force trajectory is obtained by executing the nominal skill in an environment with environmental perturbations. The modified skill model is then obtained by correcting the nominal skills to ensure that the contact force trajectory in the non-nominal environment reaches the desired state. Subsequently, the force feedback model PMDRNN is trained using the correction term $C$ of the skill model and the error $\Delta F$ between the desired contact force and the actual contact force. Finally, the trained force feedback model is utilized for real-time adjustment of DMPs to correct the force deviation in real-world robot polishing tasks.

\textbf{Contributions} Firstly, a novel force sensor feedback model called PMDRNN is proposed to address the mapping problem between multidimensional contact force errors and trajectory skill corrections in robot polishing tasks. Secondly, the combination of PMDRNN and DMPs enables real-time adjustment of the polishing skill model based on force feedback, thereby enhancing the self-adaptability of the skill model. Thirdly, real robot polishing experiments validate the effectiveness of the proposed method,  demonstrating that the inclusion of the force feedback model improves the robustness of the skill model.

\section{Related Work}

\textbf{Contact tasks skills learning from demonstration} In robotic contact tasks, the robot's position \cite{RN340}, orientation \cite{RN341}, contact force \cite{RN342}\cite{RN344} and impedance \cite{RN339}\cite{RN338}\cite{RN343} information are closely related to the skills involved. Typically, DMPs \cite{RN63}\cite{RN288} are used to learn skills related to robot position and orientation data, with orientation data commonly represented in the form of quaternions. Chang et al. \cite{RN345} proposed the contact dynamic movement primitives to learn position, orientation and force trajectories from demonstrations. They further adapted the impedance parameters online using a control policy trained by reinforcement learning (RL). Davchev et al. \cite{RN341} explored the effects of adding Gaussian perturbations in different forms when utilizing DMPs for skill learning in contact tasks. These perturbations were applied to various components, such as the forcing term, phase-modulated coupling term, and directly in the task space. Different perturbations were found to be beneficial for different tasks, for example, residual learning in the task space improved the robustness of the peg-in-hole skill model. Yu et al. \cite{RN346} used electromyography (EMG)-based method to estimate human upper limb stiffness and impedance information. They then employed a DMPs model to simultaneously capture movement and impedance features. Kim et al. \cite{RN347} proposed a neural network-based movement primitive (NNMP) to learn a continuous trajectory, which could be used as input to a force controller. In robotic polishing tasks, researchers usually use DMPs to model the position, orientation and force information of the polishing task separately \cite{RN335}\cite{RN361}. Force tracking is then implemented through a skills model with either an impedance or proportional integral differential (PID) controller. However, fine planning of the parameters for force tracking controllers is usually required, and the limited trajectory adjustment range of force tracking control is not conducive to generalizing the trajectory of robotic polishing tasks considering force feedback.

\textbf{Sensor feedback learning} To enhance the robot's adaptability to environmental perturbations, it is essential to establish a mapping between sensory space errors and action space corrections. Pastor et al. \cite{RN357} initially utilized a linear feedback model for the mapping between sensor errors and action corrections. Rai et al. \cite{RN349} employed nonlinear differential equations to represent a reactive modification term for movement plans and used a neural network to learn a reactive policy from human demonstrations. To incorporate the movement phase dependency into the feedback model, Sutanto et al. \cite{RN348} proposed phase-modulated neural networks (PMNNs), which could learn phase-dependent feedback models. Building upon this, Sutanto et al. \cite{RN327} presented a full framework for learning feedback models for reactive motion planning and used a sample-efficient RL algorithm to fine-tune these feedback models for novel tasks through a limited number of interactions with the real system. It is worth noting that all these sensor feedback models are involved in the tuning of the skill model as one term of the DMPs. Moreover, these methods are mainly applied to the tactile devices.

\section{Preliminaries}
\label{sec:preliminaries}
\subsection{DMPs in Cartesian space}

The task skills model DMPs in Cartesian space are divided into position DMPs and orientation DMPs \cite{RN334}\cite{RN63}\cite{RN288}. After collecting the position and orientation data $\{ {t_k},{p_k},{q_k}\}, {\rm{  }}k \in [1,T]$ from the demonstrated task, DMPs are used to model the related skills. Where ${t_k}$ denotes the time series, ${p_k}$ and ${q_k}$ denote the position and quaternion orientation sequences at the end of the robot in Cartesian space, respectively.

\textbf{Position DMPs:} 
\begin{equation}
  \tau \dot z = {\alpha _p}({\beta _p}({g_p} - p) - z) + {f_p}(s)
\end{equation}
\begin{equation}
  \tau \dot p = z
\end{equation}

\textbf{Orientation DMPs:}
\begin{equation}
  \tau \dot r =  - {\alpha _e}({\beta _e}{e_Q} + r) + diag(2\log ({q_g}*{\overline q _0})) + {f_q}(s)
\end{equation}
\begin{equation}
  \tau {\dot e_Q} = r
\end{equation}
\begin{equation}
  {e_Q} = 2\log ({q_g}*\overline q )
\end{equation}

Where the ${\alpha _p}$ ,${\beta _p}$ ,${\alpha _e}$ and ${\beta _e}$  are constant coefficients, and ${\beta _p} = {\alpha _p}/4$ , ${\beta _e} = {\alpha _e}/4$. ${g_p}$ and ${q_g}$ denote the robot position and orientation at the end point. And ${q _0}$ denotes the robot orientation at the starting point. ${p}$ and ${q}$ denote the position and quaternion orientation of robot movement. ${s}$ is the phase term of the canonical system. Moreover, ${z}$ and ${r}$ are intermediate variables. ${\overline q}$ is the conjugate of a unit quaternion ${q}$. The nonlinear forcing terms ${f_p}(s)$ and ${f_q}(s)$ are defined as linear combinations of \emph{M} radial basis functions ${\Psi _i}(s)$(\ref{eq:f_p})-(\ref{eq:h_i}). More details can be found in \cite{RN334} and \cite{RN288}.
\begin{equation}
  {f_p}(s) = \frac{{\sum\nolimits_{i = 1}^M {{w_{i,p}}{\Psi _i}(s)} }}{{\sum\nolimits_{i = 1}^M {{\Psi _i}(s)} }}s\label{eq:f_p}
\end{equation}
\begin{equation}
  {f_q}(s) = \frac{{\sum\nolimits_{i = 1}^M {{w_{i,q}}{\Psi _i}(s)} }}{{\sum\nolimits_{i = 1}^M {{\Psi _i}(s)} }}s\label{eq:f_q}
\end{equation}
\begin{equation}
  {\Psi _i}(s) = \exp ( - {h_i}{(s - {c_i})^2})\label{eq:psi}
\end{equation}
\begin{equation}
  {c_i} = i/M,{c_i} \in [0,1]\label{eq:c_i}
\end{equation}
\begin{equation}
  {h_i} = \frac{1}{{2{{({c_{i + 1}} - {c_i})}^2}}},({h_M} = {h_{M - 1}})\label{eq:h_i}
\end{equation}


\subsection{Modulation terms of DMPs}
In general, there are three ways to adapt the DMPs model to enhance the robustness and generalization performance of the DMPs skills model \cite{RN341}, which are adjusting the forcing term ${f_p}{(s)_{\omega  + \eta }}$, the coupling feedback term $C(\eta )$ and adjusting directly in the action space, as shown in (\ref{eq:dmpm}).

\begin{equation}
  \tau \dot z = {\alpha _p}({\beta _p}({g_p} - p) - z) + {f_p}{(s)_{\omega  + \eta }}{\rm{ + }}C(\eta ) + \eta \label{eq:dmpm}
\end{equation}

Where $\eta $ denotes the added bias. If the skill model is adjusted by the feedback term, the forcing term determines the nominal trajectory, and the phase-modulated feedback/coupling term $C(\eta )$ makes an adaptation to the skill model based on sensor feedback. In this study, the skill modulation is centered around the error in the force measured by the sensor, so the main focus is on the design of $C(\eta )$.


\section{Force Feedback Learning Model:\\Phase-Modulated Diagonal Recurrent\\Neural Networks}
\label{sec:phase-modulated diagonal recurrent neural networks}

This section focuses on the proposed feedback model PMDRNN in robot polishing tasks and the force feedback-based skill model correction method achieved through the joint implementation of PMDRNN and skill models DMPs.

\begin{figure*}[htbp] 
  \centering 
  \includegraphics[width=12cm]{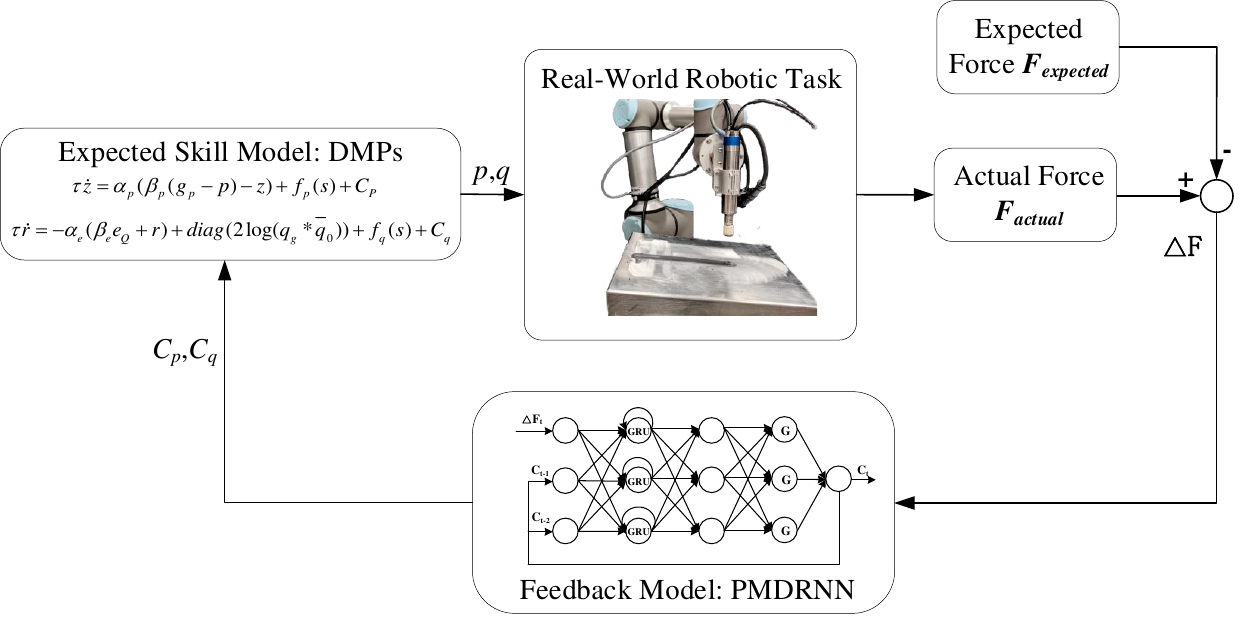} 
  \caption{Skill adjustment framework via feedback models} 
  \label{Figure2} 
\end{figure*}

A feedback model-based skill adjustment framework combining PMDRNN and DMPs is shown in Fig.\ref{Figure2}. The trained feedback model PMDRNN is utilized to fine-tune the DMPs model, enabling the adjustment of the contact force to reach the desired state in the actual environment. In the context of polishing tasks, the feedback model for contact force is employed to predict the adjustment term ${C}$ of the skill model DMPs, which is dependent on the error between the desired contact force and the actual contact force.

\begin{figure*}[htbp] 
  \centering 
  \includegraphics[width=12cm]{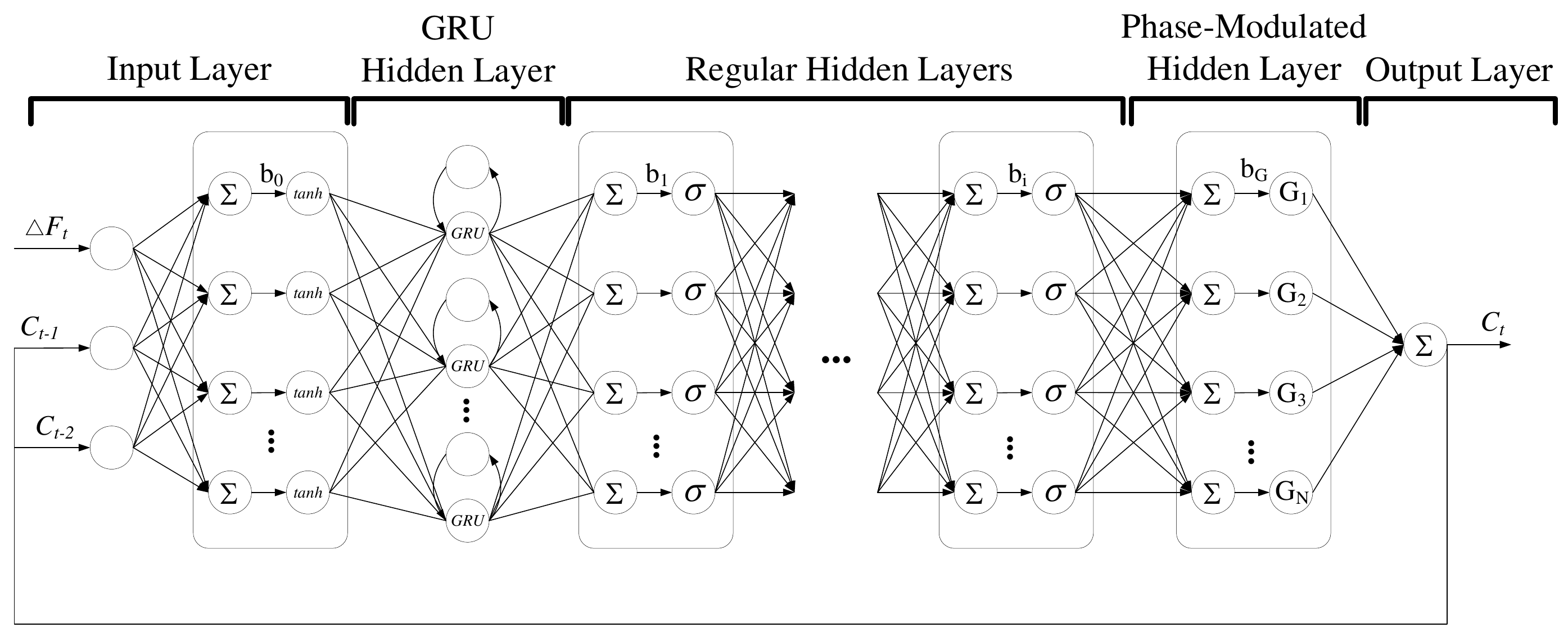} 
  \caption{The architecture of PMDRNN} 
  \label{Figure3} 
\end{figure*}

PMDRNN is a framework designed to learn force/torque feedback models from human demonstrations. Given the complexity of the contact model in robot polishing tasks and the time-series-dependent nature of contact dynamics, the DRNN \cite{RN337} structure is utilized to learn the sensor feedback model. Moreover, to address the challenges of gradient explosion and gradient disappearance, the gate recurrent unit (GRU) is used to learn time series information, as it offers computational efficiency compared to the long short-term memory (LSTM) \cite{RN356}. Furthermore, the phases of actions are incorporated into the network construction to make the feedback model dependent on the evolution of phases \cite{RN327}, enabling improved scalability of the skill model in the time domain. Fig.\ref{Figure3} depicts the architecture of the PMDRNN, which mainly consists of an input layer, a GRU hidden layer, regular hidden layers, a phase-modulated hidden layer and an output layer. The input data consists of the force sensor feedback error, as well as the outputs of the previous two moments, ${C_{t - 1}},{C_{t - 2}}$.

\textbf{Input layer:} The input to the PMDRNN are the error between the actual contact force and the expected contact force $\Delta F = {F_{actual}} - {F_{expected}} \in {\mathbb{R}^6}$, and the outputs of the previous two time steps of the PMDRNN ${C_{t - 1}},{C_{t - 2}}$. The output of the input layer is $h_{input}^t = \tanh ({W_F}\Delta {F_t} + {W_{C1}}{C_{t - 1}} + {W_{C2}}{C_{t - 2}} + {b_{input}})$. Where ${W_F},{W_{C1}}$, and ${W_{C2}}$ denote the weight matrixes between different neural network layers, ${b_{input}}$ is the bias vector of this layer. The structure of the input layer is more suitable for dynamic system learning, which is consistent with the characteristics of polishing contact.

\textbf{GRU:} The GRU hidden layer is a self-recurrent layer, which has better learning performance for sequence data and is more suitable for stable learning of sequence data with a small sample size compared to RNN and LSTM. The output of the GRU hidden layer is (\ref{eq:h_gru}).
\begin{equation}
  {r_t} = sigmoid({W_r}h_{input}^t + {U_r}h_{GRU}^{t - 1} + {b_r})
\end{equation}
\begin{equation}
  {z_t} = sigmoid({W_z}h_{input}^t + {U_z}h_{GRU}^{t - 1} + {b_z}))
\end{equation}
\begin{equation}
\mathord{\buildrel{\lower3pt\hbox{$\scriptscriptstyle\frown$}} 
\over h} _{GRU}^t = \tanh ({W_h}h_{input}^t + {U_h}({r_t} \circ h_{GRU}^{t - 1}) + {b_h})
\end{equation}
\begin{equation}
h_{GRU}^t = {z_t} \circ h_{GRU}^{t - 1} + (1 - {z_t}) \circ \mathord{\buildrel{\lower3pt\hbox{$\scriptscriptstyle\frown$}} 
\over h} _{GRU}^t\label{eq:h_gru}
\end{equation}

Where ${r_t}$ and ${z_t}$ represent the reset gate and the update gate. ${W_r}, {W_z}, {W_h}, {U_r}, {U_z}$, and ${U_h}$ denote the weight matrixes and ${b_r}, {b_z}$, and ${b_h}$ denote the bias vectors.

\textbf{Hidden layers:} The hidden layers perform nonlinear processing on the output of the previous layer to extract signal features.  The output of one of the hidden layers is: $h_i^t = sigmoid({W_{{h_i}}}h_{i - 1}^t + {b_{{h_i}}})$. Where, \emph{i} represents the index of the hidden layer, and if it is the first hidden layer, $h_{i - 1}^t = h_{GRU}^t$.

\textbf{Phase-modulated hidden layer:} The phase-modulated hidden layer considers the motion phase in this network, making the feedback model relevant to the motion phase. Its output is  defined as (\ref{eq:h_pm}).
\begin{equation}
  h_{pm}^t = G \odot ({W_{{h_{pm}}}}h_L^t + {b_{{h_{pm}}}})\label{eq:h_pm}
\end{equation}
\begin{equation}
  G = {\left[ {{G_1}{\rm{ }}{G_2}{\rm{ }} \cdots {\rm{ }}{G_N}} \right]^T}
\end{equation}
\begin{equation}
  {G_i}(s,u) = \frac{{{\psi _i}(s)}}{{\sum\nolimits_{j = 1}^N {{\psi _j}(s)} }}u{\rm{       }}, i = 1, \ldots ,N
\end{equation}

Where $h_L^t$ is the output of the last hidden layer. $s$ and $u$ are the phase variable and phase velocity, respectively. ${{\psi _i}(s)}$ denotes the basis function.

\textbf{Output Layer:} The output $C$ of the output layer is the weighted output of the phase-modulated hidden layer: $C = w_C^T{h_{pm}}$.


\section{Experiments and Analysis}
\label{sec:experiments}

In this study, the PMNN and PMDRNN are trained and compared by using real robot polishing data to evaluate their performance. The superiority of our proposed method is demonstrated through regression learning results in robotic polishing applications. Additionally, we integrate the trained force feedback model into the robot polishing skill model to assess its impact on robustness in the face of environmental perturbations.


\subsection{Experimental setup}

\begin{figure}[htbp] 
  \centering 
  \includegraphics[width=6cm]{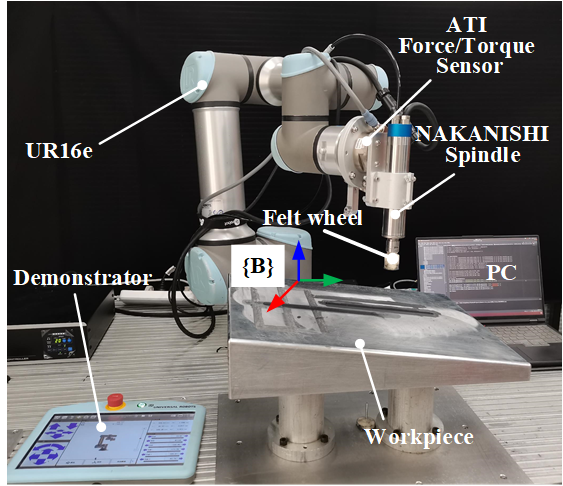} 
  \caption{The robotic polishing platform} 
  \label{Figure4} 
\end{figure}

Validation experiments of the robot polishing feedback model are conducted on a robotic polishing demonstration platform (Fig.\ref{Figure4}),  which consists of an UR16e robot, a demonstrator, an ATI Gamma force/torque sensor, a NAKANISHI spindle, a MiSUMi felt wheel, an aluminum workpiece, and a PC. The felt wheel has a diameter of 25mm and a thickness of 26mm.

\begin{figure}[htbp] 
  \centering 
  \includegraphics[width=8cm]{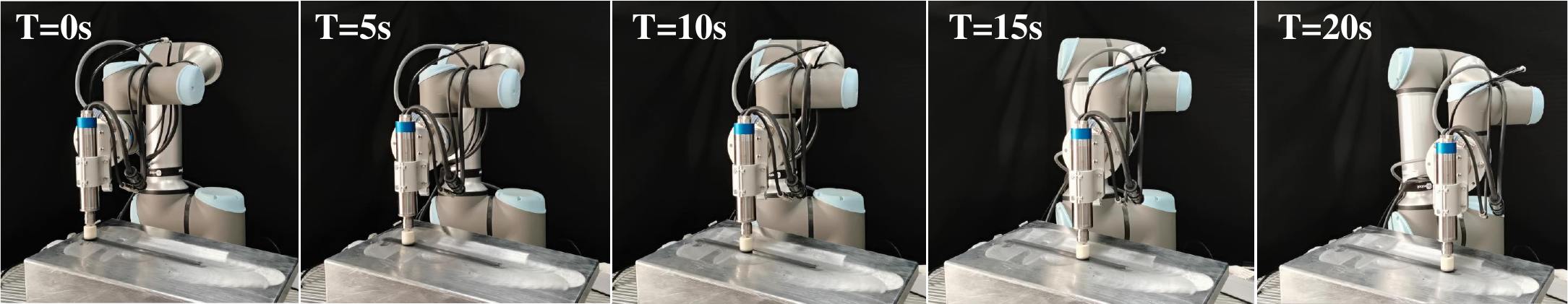} 
  \caption{Demonstrated robotic polishing trajectory} 
  \label{Figure5} 
\end{figure}

In this study, it is assumed that the contact between the felt wheel and the workpiece is non-rigid. Robot polishing demonstration experiments rely on the impedance control. Experiments with different environmental settings are conducted to collect the dataset $\left\{ {{t_k},{p_k},{q_k},{f_k}} \right\}, {\rm{  }}k \in [1,T]$ for force feedback model training. Different environments here refers to different contact forces in different tool-workpiece contact states. Specifically, different contact states and contact forces are obtained for the same skill execution due to, for example, uncertainty in the position of the workpiece with respect to the robot and the tool. ${f_k}$ is the contact force-torque in Cartesian space, and the data is sampled at 50 Hz. In this study, the spindle speed is set to 2000rpm and the specification of the polishing paste is W10-2000 mesh. In addition, a specific polishing experiment (Fig.\ref{Figure5}) is conducted to validate the proposed method, focusing on the adjustment of robot position data while keeping the robot orientation constant.

\subsection{Performance Comparison between PMNN and PMDRNN}
PMNN and PMDRNN are used to learn the data of the demonstration respectively. The total number of layers is set to 5 for both models, i.e., there are three regular hidden layers for PMNN and one regular hidden layer for PMDRNN. The error term is set as the sum of squares due to error (SSR). The learning rate is set as 0.02, the batch size is 8, and the numbers of neurons in the hidden layers are 20. The input is the contact force error between the nominal and the non-nominal demonstration experiments, and the output is the difference between the forcing term of the nominal experimental skill model and the non-nominal experimental skill model.

\begin{figure}[htpb]
\centering
\subfigure[Training results on dataset 1]{
\includegraphics[width=5.5cm]{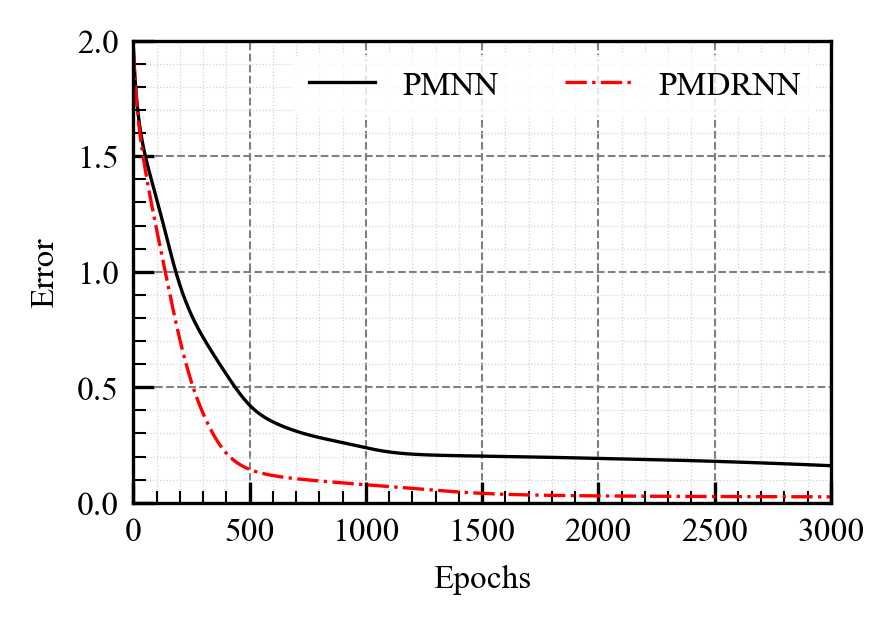}\label{1} 
}
\quad
\subfigure[Training results on dataset 2]{
\includegraphics[width=5.5cm]{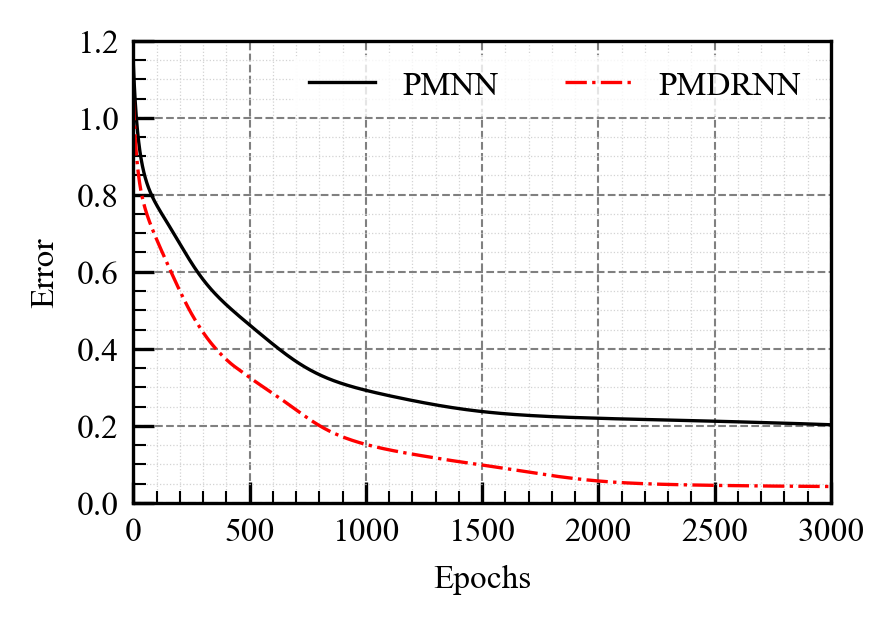} \label{2} 
}
\quad
\caption{Comparison of regression results of PMNN and PMDRNN}
\label{Figure6}
\end{figure}

In the real-world robotic polishing experiment, the training results are shown in Fig.\ref{Figure6}. Dataset 1 is obtained from experiments with different start and end points, while dataset 2 is obtained from experiments with the same start and end points. After 3000 training epochs on dataset 1, PMDRNN achieves an error of 0.025, compared to PMNN's error of 0.16, representing an 84\% error reduction. On dataset 2, PMDRNN reaches an error of 0.042, while PMNN had an error of 0.202, showing a 79\% error reduction. These results highlight PMDRNN's superior learning performance for training robot polishing task force feedback models, providing a strong basis for adjusting contact task skill models based on contact force feedback.

The superior performance of PMDRNN over PMNN in force feedback skill adjustment model learning comes mainly from the introduction of DRNN. Because the input force error term is a time-series signal and it resembles a dynamic system between the input force error term and the output skill model adjustment term, the use of DRNN can model this type of interaction data better.

\subsection{Robotic polishing experiments using PMDRNN-DMPs}

\begin{figure*}[htbp] 
  \centering 
  \includegraphics[width=15cm]{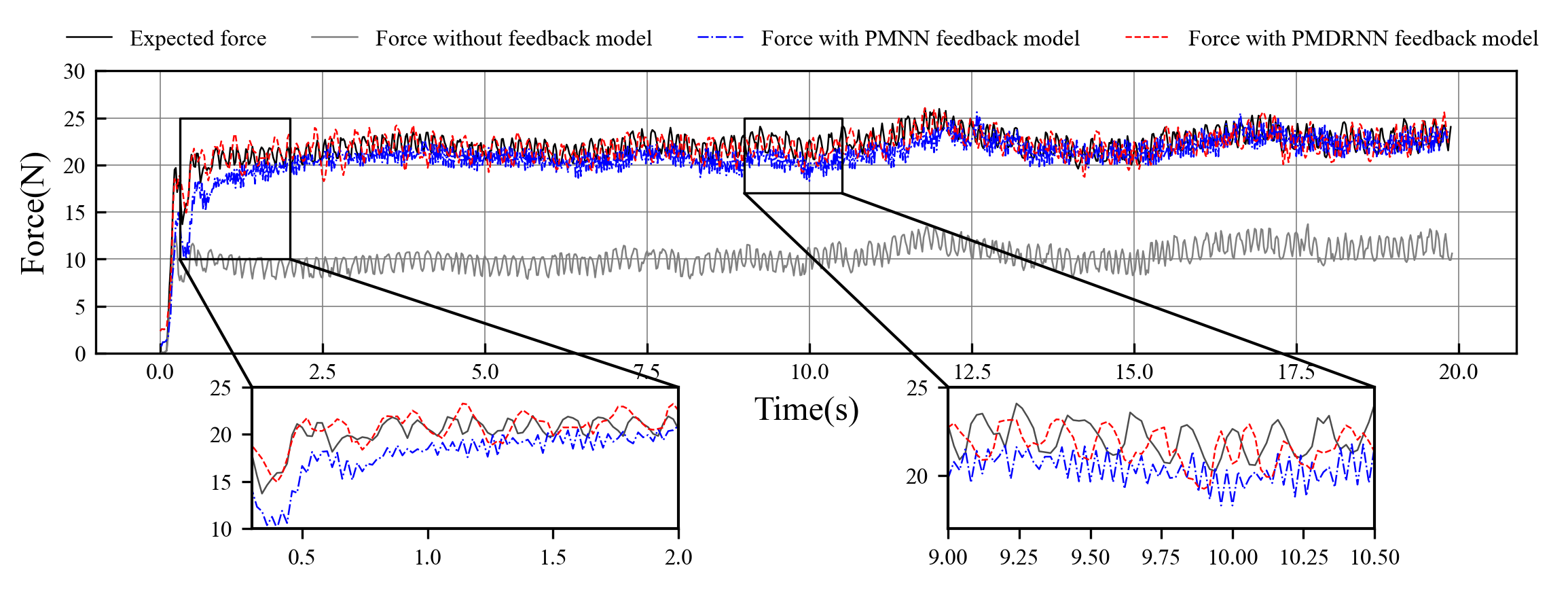} 
  \caption{Experimental results of contact force feedback skill tuning for robotic polishing} 
  \label{Figure7} 
\end{figure*}

After learning the contact force feedback phase-modulated coupling term between the desired and actual contexts, the PMDRNN adjusts the skill model online. When an undesired state occurs in the task, the skill model adapts based on feedback error to achieve the desired outcome. Experimental results (Fig. \ref{Figure7}) show that the adjusted skill model effectively tracks the desired contact force (fluctuating between 20N and 25N), with the red dashed curve closely matching it. In contrast, PMNN (blue dotted dashed line) exhibits a larger root mean square error (RMSE) of 1.83N compared to PMDRNN's 1.47N, indicating a 19.7\% improvement in force tracking accuracy for PMDRNN. This advantage is especially evident in the first half of the experiment. 12.5s ago, PMNN has an RMSE of 2.14N, while PMDRNN has an RMSE of 1.51N, representing a 29.4\% improvement in force tracking accuracy for PMDRNN. These results confirm the superiority of the proposed method, particularly in achieving closer-to-desired force tracking. Throughout real-world experiments, variations in the relative positions of the workpiece and the tool introduce inherent uncertainties into the resulting contact force. Notably, amalgamating the force error components from the preceding two moments emerges as a more advantageous strategy for refining the skill model. The convergence of performance between the two models beyond the 12.5-second mark can be attributed to the resemblance between data gathered during real-world skill-correction experiments and the training data obtained from demonstration experiments.

The effectiveness of the force feedback learning method in this study is validated in a real-world robot polishing task, and more importantly, this provides a basis for the next step of generalization between different tasks in combination with RL. 

\section{Conclusion}
\label{sec:conclusion}

This paper presents the application of a feedback learning model, PMDRNN, to robotic polishing for contact force feedback and skill model correction. In the task of learning from the robot polishing force dataset, the training accuracy of the recurrent neural network PMDRNN, which considers temporal and dynamic features, outperforms that of the feedforward neural network PMNN. This demonstrates the advantages of the proposed method. In a real-world robot polishing task, the PMDRNN and DMPs are combined to adjust the skill model online, making it closer to the requirements of the desired task, and the experimental results demonstrate the effectiveness of the adjustment strategy, providing a basis for further improving the generalization performance and robustness of the skill model by RL and other methods. The PMDRNN is a recurrent neural network, so the prediction can be made only after some time steps. And the model can only be used in non-rigid contact tasks such as polishing. More complex continuous contact tasks and more kinds of application scenarios require the use of more complex feedback models, and how to ensure the learning accuracy and computing speed of the model at the same time is a problem that needs further research.

\section{ACKNOWLEDGMENT}
The work was supported by the National Natural Science Foundation of China (Grant Nos. 52105515, U20A20294 and 52175463).

\bibliographystyle{IEEEtran.bst}
\bibliography{ref} 


\end{document}